\documentclass[11pt]{article}
\usepackage{placeins}
\usepackage{amsmath,amssymb}
\usepackage{graphicx}
\usepackage{hyperref}
\usepackage{natbib}
\usepackage{geometry}
\usepackage{amsthm}
\newtheorem{proposition}{Proposition}
\newtheorem{definition}{Definition}
\texorpdfstring

\title{\textbf{Horizon Reduction as Information Loss in Offline Reinforcement Learning}}
\author{
\textbf{Uday Kumar Nidadala} \\
Independent Researcher
\and
\textbf{Venkata Bhumika Guthi} \\
Independent Researcher
}

\date{}
\begin{document}

\maketitle

\begin{abstract}
Horizon reduction is a common design strategy in offline reinforcement learning (RL), used to mitigate long-horizon credit assignment, improve stability, and enable scalable learning through truncated rollouts, windowed training, or hierarchical decomposition \citep{levine2020offline,prudencio2023offline,park2025}. Despite recent empirical evidence that horizon reduction can improve scaling on challenging offline RL benchmarks, its theoretical implications remain underdeveloped \citep{park2025}.

In this paper, we show that horizon reduction can induce fundamental and irrecoverable information loss in offline RL. We formalize horizon reduction as learning from fixed-length trajectory segments and prove that, under this paradigm and any learning interface restricted to fixed-length trajectory segments, optimal policies may be statistically indistinguishable from suboptimal ones even with infinite data and perfect function approximation. Through a set of minimal counterexample Markov decision processes (MDPs), we identify three distinct structural failure modes: (i) prefix indistinguishability leading to identifiability failure, (ii) objective misspecification induced by truncated returns, and (iii) offline dataset support and representation aliasing.

Our results establish necessary conditions under which horizon reduction can be safe and highlight intrinsic limitations that cannot be overcome by algorithmic improvements alone, complementing algorithmic work on conservative objectives and distribution shift that addresses a different axis of offline RL difficulty \citep{fujimoto2019off,kumar2020conservative,gulcehre2020rl}.
\end{abstract}

\section{Introduction}

Offline reinforcement learning aims to learn decision-making policies from a fixed dataset of logged experience, without further interaction with the environment \citep{lange2012batch,levine2020offline,prudencio2023offline}. This setting arises naturally in high-stakes domains such as robotics, healthcare, and recommender systems, where online exploration is costly or unsafe \citep{levine2020offline,prudencio2023offline}. However, offline RL is fundamentally constrained by the information content and support of the dataset: distribution shift and extrapolation error can cause learned value functions and policies to rely on unsupported actions and transitions, leading to instability and suboptimality \citep{fujimoto2019off,kumar2020conservative,gulcehre2020rl}. These constraints become especially acute in long-horizon settings, where delayed consequences are difficult to infer from logged data and credit assignment becomes increasingly challenging \citep{arumugam2021does}.

A widely adopted response to this challenge is \emph{horizon reduction}. Modern offline RL methods frequently rely on truncated rollouts, short-horizon value targets (e.g., $n$-step returns), windowed sequence modeling, or hierarchical decompositions that optimize over fixed-length trajectory segments rather than full episodes \citep{chen2021decision,janner2021offline,levine2020offline,park2025}. These techniques have been shown to improve stability and scalability, and recent work suggests that explicitly reducing the effective decision horizon can alleviate the ``curse of horizon'' that inhibits scaling on complex tasks \citep{park2025}.

Despite these advances, existing analyses largely focus on empirical performance or algorithm specific properties. The structural consequences of horizon reduction itself independent of any particular algorithm remain underexplored \citep{levine2020offline,prudencio2023offline}. In particular, it is unclear when reducing the horizon preserves the information necessary to identify optimal behavior, and when it fundamentally destroys it. This gap is emphasized even in work advocating horizon reduction for scalability, which notes that current approaches only mitigate certain failure modes and rely on assumptions about short-segment optimality and compositionality \citep{park2025}.

In this work, we argue that horizon reduction should be viewed not merely as an approximation, but as a transformation of the learning problem that can induce irreversible information loss. We show that, in offline RL, restricting learning objectives to fixed-length trajectory segments can make optimal and suboptimal policies indistinguishable, even in simple deterministic Markov decision processes (MDPs) and even with unlimited data. This complements (rather than contradicts) the empirical observation that horizon reduction can help scaling: our results isolate when it cannot be safe in principle, regardless of algorithmic sophistication.

\paragraph{Contributions.}
Our contributions are as follows:
\begin{enumerate}
    \item We formalize horizon reduction in offline RL as learning from fixed-length trajectory segments and characterize its implications for policy identifiability \citep{levine2020offline,chen2021decision,janner2021offline}.
    \item We construct minimal counterexample MDPs demonstrating three distinct failure modes of horizon reduction: identifiability failure, objective misspecification, and support/representation aliasing.
    \item We propose a unifying notion of \emph{$H$-sufficiency}, capturing when a dataset and task admit safe horizon reduction, separating structural impossibility from classical offline RL optimization issues \citep{kumar2020conservative,gulcehre2020rl}.
    \item We discuss implications for the design and evaluation of offline RL algorithms, including conservative and pessimistic methods that address distribution shift but do not resolve horizon-induced indistinguishability \citep{fujimoto2019off,kumar2020conservative,prudencio2023offline}.
\end{enumerate}

\section{Preliminaries}

\subsection{Markov Decision Processes}

We consider finite-horizon Markov decision processes (MDPs) defined by a tuple
$M = (\mathcal{S}, \mathcal{A}, P, r, T)$, where $\mathcal{S}$ is a finite state space,
$\mathcal{A}$ is a finite action space, $P(s' \mid s, a)$ is the transition kernel,
$r(s, a, s') \in \mathbb{R}$ is the reward function, and $T$ is the episode length
\citep{sutton2018reinforcement}. A policy $\pi(a \mid s)$ induces a distribution over
trajectories $\tau = (s_0, a_0, s_1, \ldots, s_T)$, with return
\begin{equation}
G(\tau) = \sum_{t=0}^{T-1} r_t .
\end{equation}

\subsection{Offline Reinforcement Learning}

In offline reinforcement learning, the learner is given access only to a fixed dataset
$\mathcal{D}$ of trajectories generated by an unknown behavior policy $\mu$. The learner’s
goal is to compute a policy $\pi$ maximizing expected return under the true MDP, using
only $\mathcal{D}$ \citep{lange2012batch,levine2020offline,prudencio2023offline}.

A central difficulty in offline RL is that the dataset may not contain sufficient
information to evaluate or compare policies, particularly when policy differences
manifest only through long-range effects. This interacts with distribution shift and
extrapolation error, motivating conservative and behavior-regularized approaches such as
batch-constrained learning and conservative Q-learning
\citep{fujimoto2019off,kumar2020conservative}, as well as broader pessimism-based
perspectives on offline RL \citep{rashidinejad2021bridging,prudencio2023offline}.

\subsection{Horizon Reduction}

We use the term \emph{horizon reduction} to refer to any learning procedure that replaces
the full-horizon objective
\begin{equation}
J(\pi) = \mathbb{E}_{\tau \sim \pi} \left[ \sum_{t=0}^{T-1} r_t \right]
\end{equation}
with an objective that depends only on fixed-length trajectory segments of length
$H < T$.

Examples include truncated returns and short-horizon value targets
\citep{daley2024truncated,park2025}, windowed sequence modeling
\citep{chen2021decision,janner2021offline}, and hierarchical decompositions that reduce
the effective policy horizon \citep{sutton1999between,park2025}.

Formally, such methods rely only on statistics of sub-trajectories
$\tau_{t:t+H} = (s_t, a_t, \ldots, s_{t+H})$, marginalizing over the remainder of the
episode. This formalization directly matches the training interface used in many
``windowed'' offline pipelines, where the learning algorithm is applied to fixed-length
segments rather than full episodes \citep{chen2021decision,janner2021offline}.

\section{Horizon Reduction as Information Loss}

We now present the central conceptual claim of this paper: horizon reduction can destroy
information necessary to identify optimal policies in offline reinforcement learning.

The key observation is that two policies may induce identical distributions over all
length-$H$ trajectory segments observable to a learner whose decision rule depends only
on such segments, while inducing different full-horizon returns. In such cases, no
learning algorithm whose decision rule depends only on these segments can reliably prefer
the better policy. This is not a limitation of function approximation, optimization, or
sample size. Rather, it is an information-theoretic impossibility arising from the
restriction to truncated segments.

This perspective is conceptually aligned with the broader view that long-horizon credit
assignment can be limited by information availability rather than only reward sparsity
\citep{arumugam2021does}.

In the remainder of this section, we formalize this claim and illustrate it through
concrete counterexamples.

\section{Counterexamples}

In this section, we construct three minimal Markov decision processes (MDPs)
demonstrating distinct failure modes of horizon reduction in offline reinforcement
learning. Each example is deterministic, finite, and small, yet exhibits irrecoverable
failure under horizon reduction.

\subsection{Prefix-Indistinguishable Commitment}

We first show that horizon reduction can destroy policy identifiability even in a
deterministic MDP with sparse rewards. The failure arises because the consequences of the
initial action are revealed only beyond the reduced horizon, while all shorter trajectory
prefixes are identical. This resembles the structural risk of windowed training: if the
window excludes decision-relevant commitments, the learning problem becomes non-identifiable
regardless of estimator quality \citep{chen2021decision,janner2021offline}.

\begin{figure}[t]
    \centering
    \includegraphics[width=0.8\linewidth]{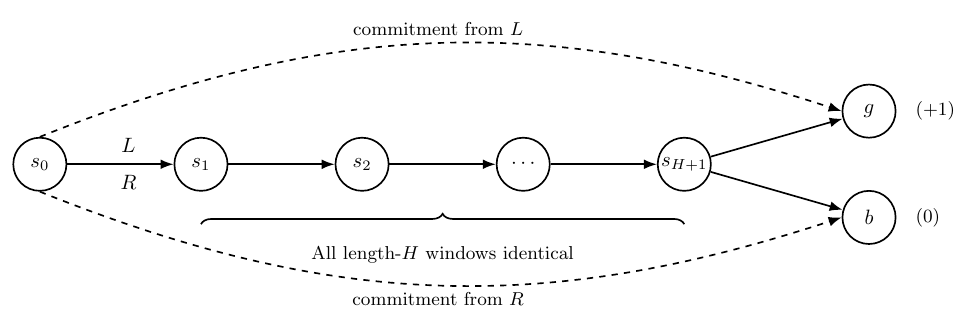}
    \caption{Prefix-indistinguishable commitment. Both initial actions $(L)$ and $(R)$
    lead to identical length-$H$ trajectory windows along the shared chain. The value
    difference induced by the initial action is only revealed at step $H+1$, outside the
    reduced horizon.}
    \label{fig:prefix}
\end{figure}

\paragraph{MDP construction.}
Let the episode length be $T = H + 2$. The state space is
\[
\mathcal{S} = \{ s_0, s_1, s_2, \ldots, s_{H+1}, g, b \}.
\]

At the initial state $s_0$, the agent chooses between two actions
$\mathcal{A}(s_0) = \{L, R\}$. At all subsequent chain states
$s_1, \ldots, s_{H+1}$, only a single action is available. Both actions
$L$ and $R$ deterministically transition from $s_0$ to $s_1$. The chain then
progresses deterministically according to
\[
s_t \rightarrow s_{t+1}, \quad t = 1, \ldots, H.
\]

At the final step, the terminal outcome depends on the initial action: choosing
$L$ causes the trajectory to transition from $s_{H+1}$ to terminal state $g$
with reward $1$, while choosing $R$ transitions to terminal state $b$ with reward
$0$. All intermediate rewards are zero.

To maintain a proper Markov formulation, the commitment induced by the initial
action can be encoded into an augmented state variable that is not observed
within length-$H$ trajectory windows (e.g., using distinct states
$s_t^L$ and $s_t^R$). We assume, however, that the learning procedure’s decision
rule is restricted to fixed-length trajectory windows that exclude the initial
action, as is standard in cropped or windowed offline training
\citep{chen2021decision,janner2021offline}.

\paragraph{Offline dataset.}
Consider an offline dataset generated by a behavior policy that selects $L$ and
$R$ with positive probability at $s_0$. The learning procedure operates exclusively
on length-$H$ trajectory segments that begin at $s_1$, as is common in windowed or
truncated training \citep{chen2021decision,janner2021offline}. Under this
construction, every observable length-$H$ segment is identical, regardless of
whether the initial action was $L$ or $R$.

\begin{proposition}[Prefix indistinguishability]
For the MDP described above and any learning algorithm whose decision rule depends
only on the distribution of length-$H$ trajectory segments in the dataset, the
actions $L$ and $R$ at $s_0$ are statistically indistinguishable. Consequently, no
such algorithm can be guaranteed to select the optimal action $L$, even with
infinite data and perfect function approximation.
\end{proposition}

\paragraph{Proof sketch.}
Under the dataset construction, the conditional distribution over all length-$H$
segments observable to the learner is identical under $L$ and $R$. Any objective or
estimator based solely on these segments therefore assigns equal value to both
actions. However, the true full-horizon returns differ: choosing $L$ yields return
$1$, while choosing $R$ yields return $0$. This failure is information-theoretic and
cannot be resolved without access to longer-horizon information.

\subsection{Short-Segment Optimality Violation}

We next show that truncating the objective can induce a policy that is optimal under the
reduced horizon but arbitrarily suboptimal under the true objective. This aligns with the
general caution that shorter-horizon targets trade off bias, variance, and objective
alignment \citep{sutton2018reinforcement,daley2024truncated}, and it directly complements
empirical work that treats value-horizon reduction (e.g., $n$-step returns) as a principal
lever for scaling \citep{park2025}.

\begin{figure}[t]
    \centering
    \includegraphics[width=0.85\linewidth]{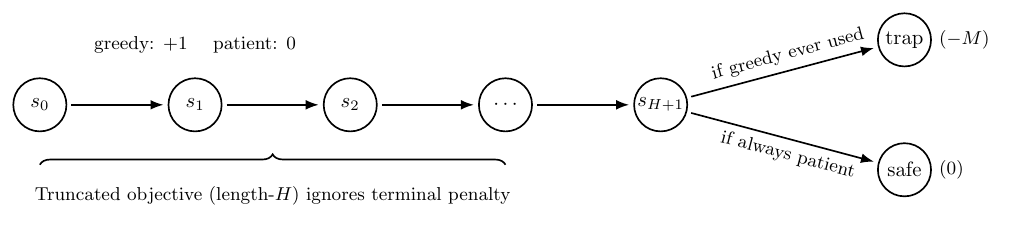}
    \caption{Short-segment optimality violation. Greedy actions yield higher immediate
    rewards on every length-$H$ segment but incur a delayed penalty beyond the reduced
    horizon. Optimizing the truncated objective selects a policy that is globally dominated
    under the true return.}
    \label{fig:shortsegment}
\end{figure}

\paragraph{MDP construction.}
Let the episode length be $T = H + 3$. The state space consists of a chain
\[
\mathcal{S} = \{ s_0, s_1, \ldots, s_H, s_{H+1}, \texttt{trap}, \texttt{safe} \}.
\]
At each state $s_t$ for $t = 0, \ldots, H$, the agent chooses between two actions
\[
\mathcal{A}(s_t) = \{ a^{\mathrm{greedy}}, a^{\mathrm{patient}} \}.
\]
From each $s_t$, both actions transition deterministically to $s_{t+1}$. However, choosing
$a^{\mathrm{greedy}}$ at any time sets an augmented-state flag $G=1$. Choosing
$a^{\mathrm{patient}}$ leaves the flag unchanged. At state $s_{H+1}$, the terminal
transition depends on this flag: if $G=1$, transition to terminal state \texttt{trap};
otherwise, transition to terminal state \texttt{safe}.

\paragraph{Rewards.}
For each $t = 0, \ldots, H$,
\[
r(s_t, a^{\mathrm{greedy}}) = +1,
\qquad
r(s_t, a^{\mathrm{patient}}) = 0 .
\]
At termination,
\[
r(\texttt{trap}) = -M,
\qquad
r(\texttt{safe}) = 0,
\]
where $M > H+1$ is a constant.

\paragraph{Offline dataset.}
Assume the offline dataset is generated by a behavior policy that heavily favors
$a_{\mathrm{greedy}}$, so that most observed trajectory segments contain immediate positive
rewards. A horizon-reduced learner evaluates policies using purely truncated returns of
length $H$, without access to rewards beyond this horizon, consistent with common
short-horizon target construction in deep RL and with value-horizon reduction as studied in
recent scaling work \citep{sutton2018reinforcement,daley2024truncated,park2025}.

\begin{proposition}[Short-segment optimality violation]
There exists a policy $\pi_{\mathrm{greedy}}$ that is optimal under the truncated objective
\begin{equation}
J_H(\pi) = \mathbb{E}\left[\sum_{t=0}^{H} r_t\right],
\end{equation}
but arbitrarily suboptimal under the true objective
\begin{equation}
J(\pi) = \mathbb{E}\left[\sum_{t=0}^{T-1} r_t\right].
\end{equation}
\end{proposition}

\paragraph{Proof sketch.}
Under $J_H$, selecting $a^{\mathrm{greedy}}$ at each step strictly dominates
$a^{\mathrm{patient}}$, yielding return $H+1$. Under the true objective, however, any
policy that selects $a^{\mathrm{greedy}}$ incurs the terminal penalty $-M$, resulting in
total return $(H+1) - M < 0$. In contrast, the patient policy achieves total return $0$.
Since $M$ can be chosen arbitrarily large, the suboptimality gap induced by horizon
truncation can be made arbitrarily severe.

This counterexample highlights that horizon reduction can change the optimization problem
itself, rather than merely introducing estimation noise an issue complementary to, but
distinct from, distribution-shift-driven failure modes typically targeted by conservative
or pessimistic offline RL objectives \citep{fujimoto2019off,kumar2020conservative,rashidinejad2021bridging}.

\subsection{Offline Support and Representation Aliasing}

Finally, we demonstrate that offline dataset support and representation choices can render
long-term distinctions unobservable under horizon reduction, even in fully observable
MDPs. This failure mode connects two widely recognized themes: (i) offline reinforcement
learning depends critically on dataset support \citep{levine2020offline,prudencio2023offline},
and (ii) representations and abstractions can collapse decision-relevant distinctions,
potentially inducing effective partial observability in the learned representation
\citep{allen2021learning,ghosh2021inducing}.

\begin{figure}[t]
    \centering
    \includegraphics[width=0.85\linewidth]{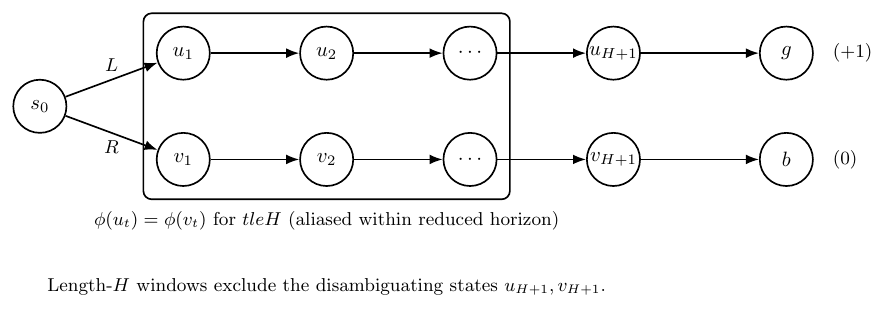}
    \caption{Offline support and representation aliasing. Although distinct long-term
    trajectories exist in the MDP, representation aliasing and windowed offline training
    collapse the two branches within the reduced horizon. Length-$H$ trajectory segments
    exclude the disambiguating states, rendering the optimal action unidentifiable.}
    \label{fig:aliasing}
\end{figure}

\paragraph{MDP construction.}
Let the episode length be $T = H + 2$. The state space consists of two deterministic chains
branching from the initial state:
\[
\mathcal{S} =
\{ s_0, u_1, \ldots, u_{H+1}, v_1, \ldots, v_{H+1}, g, b \}.
\]

At the initial state $s_0$, the agent chooses between two actions $\{L, R\}$. Action $L$
deterministically transitions to $u_1$, while action $R$ transitions to $v_1$. Each branch
then progresses deterministically according to
\[
u_t \rightarrow u_{t+1}, \qquad
v_t \rightarrow v_{t+1}, \qquad
t = 1, \ldots, H .
\]

At the final step, the branches terminate differently: state $u_{H+1}$ transitions to $g$
with reward $1$, while state $v_{H+1}$ transitions to $b$ with reward $0$. All intermediate
rewards are zero. Under the true MDP dynamics, action $L$ is strictly optimal.

\paragraph{Offline dataset and representation.}
Assume the offline dataset consists exclusively of cropped trajectory windows of length
$H$ that begin after the initial action, as in windowed or segment-based offline training
\citep{chen2021decision,janner2021offline}. Moreover, suppose the state representation used
for learning maps the branch states to a shared feature space such that
\[
\phi(u_t) = \phi(v_t), \qquad \text{for all } t = 1, \ldots, H .
\]

The disambiguating states $u_{H+1}$ and $v_{H+1}$, which reveal the value difference between
actions $L$ and $R$, never appear in any length-$H$ window in the dataset.

This setup captures realistic offline RL scenarios in which representation learning,
preprocessing, or abstraction collapses distinct latent states, and training is performed
on fixed-length segments. Such collapse can break Markov properties of the induced
representation and create history-dependent dynamics at the abstract level, undermining
identifiability from local windows \citep{allen2021learning,ghosh2021inducing}.

\begin{proposition}[Offline support-induced indistinguishability]
Under the dataset and representation described above, the actions $L$ and $R$ at $s_0$
induce identical distributions over all observable length-$H$ trajectory segments.
Consequently, any learning algorithm whose decision rule depends only on these segments
cannot distinguish the optimal action $L$ from the suboptimal action $R$.
\end{proposition}

\paragraph{Proof sketch.}
By construction, for all time steps $t \le H$, the observed representations of states along
the two branches are identical, and no length-$H$ segment contains either terminal state.
Therefore, the induced distributions over observable trajectory segments are identical
under actions $L$ and $R$. However, the true full-horizon returns differ. The failure is due
to lack of dataset support and representational aliasing and cannot be resolved without
access to longer-horizon information or richer representations.

\section{H-Sufficiency and Necessary Conditions}

The counterexamples in Section~4 demonstrate that horizon reduction can fail for
qualitatively different reasons: loss of identifiability (Section~4.1), objective
misspecification (Section~4.2), and offline support and representation limitations
(Section~4.3). In this section, we introduce a unifying notion that characterizes when
horizon reduction can be safe, and when such failures are unavoidable.

\subsection{H-Sufficiency}

We formalize the intuition that fixed-length trajectory segments must contain sufficient
information to identify optimal behavior in offline reinforcement learning.

\begin{definition}[$H$-sufficiency]
An offline reinforcement learning problem $(M, \mathcal{D})$ is said to be
$H$-sufficient if the optimal policy $\pi^\star$ can be identified from the collection of
all statistics of length-$H$ trajectory segments that a horizon-reduced learning algorithm
is permitted to depend on under dataset $\mathcal{D}$. That is, for any policy $\pi$,
\begin{equation}
\mathcal{L}_{\mathcal{D}}(\tau_{t:t+H} \mid \pi)
=
\mathcal{L}_{\mathcal{D}}(\tau_{t:t+H} \mid \pi^\star)
\quad \forall t
\;\;\Longrightarrow\;\;
J(\pi) = J(\pi^\star).
\end{equation}
\end{definition}

Intuitively, $H$-sufficiency requires that no two policies with different full-horizon
values induce identical distributions over all observable length-$H$ trajectory segments.
When this condition fails, horizon-reduced learning objectives cannot reliably distinguish
optimal from suboptimal behavior.

\subsection{Violations of $H$-Sufficiency}

Each counterexample in Section~4 violates $H$-sufficiency for a different structural
reason.

\paragraph{Prefix indistinguishability (Section~4.1).}
In the prefix-indistinguishable commitment MDP, the initial action determines the terminal
outcome, but all length-$H$ trajectory segments lie entirely within a shared prefix. As a
result, the distributions over observable segments are identical despite differing
returns. This constitutes a pure identifiability failure induced by cropped or windowed
training interfaces \citep{chen2021decision,janner2021offline}.

\paragraph{Objective misspecification (Section~4.2).}
In the short-segment optimality violation MDP, policies are distinguishable under truncated
returns, but the truncated objective $J_H$ is not aligned with the true objective $J$.
Here, $H$-sufficiency fails because optimizing over length-$H$ segments implicitly changes
the task definition. This is structurally related to bias--variance trade-offs in
multi-step targets \citep{sutton2018reinforcement,daley2024truncated} and to the empirical
use of value-horizon reduction as a scalability lever \citep{park2025}.

\paragraph{Offline support and representation aliasing (Section~4.3).}
In the third MDP, the environment itself is fully observable and Markov, but the offline
dataset and representation collapse distinct long-term trajectories into identical
observed segments. Even though disambiguating information exists in principle, it is not
supported by the dataset within the reduced horizon. This violation arises from the
interaction between horizon reduction and offline data limitations
\citep{levine2020offline,prudencio2023offline} and is consistent with theory showing that
abstraction and representation choices can induce effective partial observability and
history dependence \citep{allen2021learning,ghosh2021inducing}.

\subsection{Necessary Conditions for Safe Horizon Reduction}

The preceding analysis suggests that horizon reduction can be safe only under strong
conditions. In particular, at least one of the following must hold:
\begin{itemize}
    \item \textbf{$H$-identifiability:} all policies with different full-horizon values
    induce distinguishable distributions over length-$H$ trajectory segments observable
    in the dataset.
    \item \textbf{Objective consistency:} optimizing truncated returns yields the same
    policy ordering as optimizing full-horizon returns
    \citep{sutton2018reinforcement,daley2024truncated}.
    \item \textbf{Dataset and representation sufficiency:} the offline dataset and state
    representation preserve all decision-relevant distinctions within the reduced horizon
    \citep{levine2020offline,allen2021learning,prudencio2023offline}.
\end{itemize}

Absent such conditions, failures induced by horizon reduction are not artifacts of limited
data, function approximation, or optimization, but are intrinsic to the problem
formulation.

\section{Implications for Offline RL Design}

Our analysis shows that horizon reduction in offline reinforcement learning is not merely
a heuristic approximation, but a structural modification of the learning problem that can
fundamentally limit what is learnable from data. We now discuss the implications of these
findings for algorithm design, evaluation, and theory.

\subsection{Horizon Reduction Is Not Universally Safe}

A central implication of Sections~4 and~5 is that horizon reduction is not universally
safe, even in deterministic, fully observable MDPs with unlimited data. Failures can arise
from indistinguishability, objective misspecification, or offline dataset support
limitations. Importantly, these failures persist regardless of model capacity, dataset
size, or optimization quality.

As a result, empirical success of horizon reduction on benchmark tasks should not be
interpreted as evidence of universal correctness. Instead, such success implicitly relies
on unverified assumptions about task structure, dataset coverage, or representational
sufficiency \citep{levine2020offline,prudencio2023offline,park2025}.

\subsection{Horizon Reduction Encodes Implicit Assumptions}

Horizon-reduced objectives implicitly assume some form of $H$-sufficiency. For example:
\begin{itemize}
    \item \textbf{Truncated return objectives} assume that delayed consequences beyond
    horizon $H$ do not affect optimal decision-making, or that the resulting bias remains
    aligned with the true objective
    \citep{sutton2018reinforcement,daley2024truncated,park2025}.
    \item \textbf{Windowed or segment-based learning} assumes that decision-relevant
    information is contained within local trajectory windows, which is central to offline
    sequence modeling formulations \citep{chen2021decision,janner2021offline}.
    \item \textbf{Hierarchical decompositions} assume that short-horizon sub-policies
    compose into optimal long-horizon behavior, an assumption underlying options and
    temporally extended actions as well as modern policy-horizon reduction methods
    \citep{sutton1999between,park2025}.
\end{itemize}

Our counterexamples show that when these assumptions fail, horizon reduction induces
irrecoverable information loss.

\subsection{Offline RL Requires Dataset-Aware Reasoning}

Unlike online reinforcement learning, offline RL is constrained not only by the
environment but also by the support of the dataset. Horizon reduction can interact with
dataset limitations in subtle ways, particularly when training is performed on cropped
windows or learned representations collapse distinct long-term trajectories
\citep{levine2020offline,fujimoto2019off,kumar2020conservative,prudencio2023offline}.

This suggests that offline RL algorithms should be evaluated not only by their learning
objectives, but also by whether their data preprocessing, representation learning, and
horizon reduction choices preserve decision-relevant distinctions
\citep{allen2021learning,ghosh2021inducing}.

\subsection{Implications for Empirical Evaluation}

Our results suggest that benchmarking horizon-reduced methods solely on performance
metrics may be insufficient. In particular:
\begin{itemize}
    \item Performance improvements under horizon reduction do not guarantee policy
    correctness under the true objective.
    \item Scaling curves alone cannot reveal information-theoretic failures.
    \item Empirical success may reflect benign task structure rather than algorithmic
    robustness \citep{prudencio2023offline,park2025}.
\end{itemize}

We therefore advocate complementary evaluations that probe identifiability, long-horizon
credit assignment, and dataset sufficiency.

\subsection{Toward Principled Use of Horizon Reduction}

While our results identify fundamental limitations, they do not argue against horizon
reduction per se. Instead, they highlight the need for principled criteria governing its
use. Potential directions include:
\begin{itemize}
    \item Explicitly characterizing task classes that satisfy $H$-sufficiency.
    \item Designing diagnostics to detect horizon-induced indistinguishability.
    \item Incorporating mechanisms that preserve or recover long-range information,
    especially in windowed training regimes \citep{chen2021decision,janner2021offline}.
\end{itemize}

Absent such guarantees, horizon reduction should be viewed as a conditional strategy whose
validity depends on properties of the task and dataset.

\section{Related Work}

\paragraph{Offline Reinforcement Learning.}
Offline reinforcement learning studies policy learning from fixed datasets without
additional environment interaction \citep{lange2012batch,levine2020offline,prudencio2023offline}.
A large body of work has focused on addressing distribution shift, extrapolation error, and
dataset support mismatch through conservative objectives, pessimism, or policy constraints
\citep{fujimoto2019off,kumar2020conservative,rashidinejad2021bridging,gulcehre2020rl}.
While these approaches improve empirical robustness, they primarily address optimization
and estimation challenges under support mismatch rather than structural limits imposed by
restricting the learning objective to short segments. Our work is complementary: we
identify conditions under which optimal behavior is not identifiable from the available
information once horizon reduction is applied, independent of algorithmic choices.

\paragraph{Horizon Reduction and Truncated Objectives.}
Reducing the effective planning horizon has long been used as a practical strategy in
reinforcement learning through truncated returns, bootstrapped value targets, and
short-horizon planning \citep{sutton2018reinforcement,daley2024truncated}. Recent work
argues that horizon reduction can substantially improve scalability and stability and
empirically demonstrates strong performance gains on challenging benchmarks by reducing
both value horizon and policy horizon \citep{park2025}. These methods typically acknowledge
that horizon reduction relies on assumptions about task structure or short-horizon
optimality. Our contribution is to formalize the limits of these assumptions by constructing
counterexamples in which horizon reduction induces irrecoverable information loss or
objective misspecification.

\paragraph{Hierarchical and Segment-Based Reinforcement Learning.}
Hierarchical reinforcement learning decomposes long-horizon problems into shorter-horizon
subproblems, often via temporally extended actions (options) and skill learning
\citep{sutton1999between}. Similarly, segment-based and sequence-modeling approaches
optimize policies using fixed-length trajectory windows
\citep{chen2021decision,janner2021offline}. While these methods have proven effective in
practice, their correctness generally depends on compositionality and sufficiency
assumptions that are rarely stated explicitly. Our analysis shows that, in the absence of
such assumptions, optimizing over short segments may fail to preserve optimal long-horizon
behavior in offline settings.

\paragraph{Credit Assignment and Long-Horizon Dependencies.}
Long-horizon credit assignment has been recognized as a fundamental challenge in
reinforcement learning. Prior work highlights that difficulty can arise from limited
information flow rather than only reward sparsity, motivating information-theoretic and
structural perspectives on credit assignment \citep{arumugam2021does}. In contrast to
algorithmic mechanisms for improved credit propagation, our results show that when
learning is restricted to fixed-length segments, certain long-range dependencies can be
unidentifiable altogether.

\paragraph{Positioning of This Work.}
In summary, our work does not propose a new offline reinforcement learning algorithm, nor
does it challenge the empirical successes of horizon-reduced methods \citep{park2025}.
Instead, it provides a complementary, theory-informed perspective clarifying when horizon
reduction can be expected to work and when it cannot. By identifying necessary conditions
for safe horizon reduction, we aim to inform the design and evaluation of future offline RL
methods.

\section{Conclusion}

We have shown that horizon reduction in offline reinforcement learning can induce
fundamental information-theoretic and objective-level failures that prevent the
identification of optimal policies. By isolating three distinct structural failure modes,
we provide a principled explanation for when and why horizon reduction breaks down.

Our analysis highlights the importance of understanding the informational consequences of
design choices in offline reinforcement learning and motivates future work on identifying
tasks and dataset classes for which horizon reduction is probably safe. This complements
recent empirical evidence that horizon reduction can improve scalability, while leaving
open important theoretical questions \citep{park2025}.

\bibliographystyle{plainnat}
\bibliography{references}

@article{allen2021learning,
  title={Learning Markov state abstractions for deep reinforcement learning},
  author={Allen, Cameron and Parikh, Neil and Gottesman, Omer and Konidaris, George},
  journal={Advances in Neural Information Processing Systems},
  volume={34},
  pages={8229--8241},
  year={2021}
}

@article{arumugam2021does,
  title={An information-theoretic perspective on credit assignment in reinforcement learning},
  author={Arumugam, Dilip and Henderson, Peter and Bacon, Pierre-Luc},
  journal={arXiv preprint arXiv:2103.06224},
  year={2021}
}

@article{chen2021decision,
  title={Decision transformer: Reinforcement learning via sequence modeling},
  author={Chen, Lili and Lu, Kevin and Rajeswaran, Aravind and Lee, Kimin and Grover, Aditya and Laskin, Michael and Abbeel, Pieter and Srinivas, Aravind and Mordatch, Igor},
  journal={Advances in Neural Information Processing Systems},
  volume={34},
  pages={15084--15097},
  year={2021}
}

@article{daley2024truncated,
  title={Averaging $n$-step returns reduces variance in reinforcement learning},
  author={Daley, Benjamin and White, Martha and Machado, Marlos C.},
  journal={arXiv preprint arXiv:2402.03903},
  year={2024}
}

@inproceedings{fujimoto2019off,
  title={Off-policy deep reinforcement learning without exploration},
  author={Fujimoto, Scott and Meger, David and Precup, Doina},
  booktitle={International Conference on Machine Learning},
  pages={2052--2062},
  year={2019},
  publisher={PMLR}
}

@article{gulcehre2020rl,
  title={Addressing extrapolation error in deep offline reinforcement learning},
  author={Gulcehre, Caglar and Colmenarejo, Sergio Gomez and Sygnowski, Jakub and Paine, Thomas and Zolna, Konrad and Chen, Yunpeng and Hoffman, Matthew and Pascanu, Razvan and de Freitas, Nando},
  journal={NeurIPS 2020 Offline Reinforcement Learning Workshop},
  year={2020}
}

@article{janner2021offline,
  title={Offline reinforcement learning as one big sequence modeling problem},
  author={Janner, Michael and Li, Qiyang and Levine, Sergey},
  journal={Advances in Neural Information Processing Systems},
  volume={34},
  pages={1273--1286},
  year={2021}
}

@article{kumar2020conservative,
  title={Conservative Q-learning for offline reinforcement learning},
  author={Kumar, Aviral and Zhou, Aurick and Tucker, George and Levine, Sergey},
  journal={Advances in Neural Information Processing Systems},
  volume={33},
  pages={1179--1191},
  year={2020}
}

@incollection{lange2012batch,
  title={Batch reinforcement learning},
  author={Lange, Sascha and Gabel, Thomas and Riedmiller, Martin},
  booktitle={Reinforcement Learning: State-of-the-Art},
  pages={45--73},
  year={2012},
  publisher={Springer}
}

@article{levine2020offline,
  title={Offline reinforcement learning: Tutorial, review, and perspectives on open problems},
  author={Levine, Sergey and Kumar, Aviral and Tucker, George and Fu, Justin},
  journal={arXiv preprint arXiv:2005.01643},
  year={2020}
}

@article{park2025,
  title={Horizon reduction makes reinforcement learning scalable},
  author={Park, Sehoon and Frans, Kevin and Mann, David and Eysenbach, Benjamin and Kumar, Aviral and Levine, Sergey},
  journal={arXiv preprint arXiv:2506.04168},
  year={2025}
}

@article{prudencio2023offline,
  title={A survey on offline reinforcement learning: Taxonomy, review, and open problems},
  author={Prudencio, Ricardo F. and Maximo, Marcos R. and Colombini, Esther L.},
  journal={IEEE Transactions on Neural Networks and Learning Systems},
  volume={35},
  number={8},
  pages={10237--10257},
  year={2023}
}

@article{rashidinejad2021bridging,
  title={Bridging offline reinforcement learning and imitation learning: A tale of pessimism},
  author={Rashidinejad, Parham and Zhu, Banghua and Ma, Chao and Jiao, Jiantao and Russell, Stuart},
  journal={Advances in Neural Information Processing Systems},
  volume={34},
  pages={11702--11716},
  year={2021}
}

@article{ghosh2021inducing,
  title={Why generalization in reinforcement learning is difficult: Epistemic {POMDPs} and implicit partial observability},
  author={Ghosh, Dibya and Rahme, Jiaming and Kumar, Aviral and Zhang, Amy and Adams, Ryan P. and Levine, Sergey},
  journal={Advances in Neural Information Processing Systems},
  volume={34},
  pages={25502--25515},
  year={2021}
}

@article{sutton1999between,
  title={Between {MDPs} and semi-{MDPs}: A framework for temporal abstraction in reinforcement learning},
  author={Sutton, Richard S. and Precup, Doina and Singh, Satinder},
  journal={Artificial Intelligence},
  volume={112},
  number={1--2},
  pages={181--211},
  year={1999}
}

@book{sutton2018reinforcement,
  title={Reinforcement Learning: An Introduction},
  author={Sutton, Richard S. and Barto, Andrew G.},
  edition={2},
  publisher={MIT Press},
  address={Cambridge, MA},
  year={2018}
}
\end{document}